\documentclass[sigconf]{acmart}

\AtBeginDocument{%
  \providecommand\BibTeX{{%
    \normalfont B\kern-0.5em{\scshape i\kern-0.25em b}\kern-0.8em\TeX}}}

\copyrightyear{2025}
\acmYear{2025}
\setcopyright{acmlicensed}
\acmConference[WSDM '25] {Proceedings of the Eighteenth ACM International Conference on Web Search and Data Mining}{March 10--14, 2025}{Hannover, Germany.}
\acmBooktitle{Proceedings of the Eighteenth ACM International Conference on Web Search and Data Mining (WSDM '25), March 10--14, 2025, Hannover, Germany}
\acmISBN{979-8-4007-1329-3/25/03}
\acmDOI{10.1145/3701551.3703559}

\usepackage{amsmath,scalerel}
\usepackage{enumitem}
\usepackage{booktabs}
\usepackage{multirow}
\usepackage[ruled,linesnumbered]{algorithm2e}
\usepackage{algpseudocode}
\usepackage{color}
\usepackage{subfigure}
\usepackage{balance}

\newenvironment{ack}{\section*{ACKNOWLEDGMENTS}}{}
\newenvironment{ethic}{\section*{ETHICAL CONSIDERATIONS}}{}

\usepackage[toc,title]{appendix}
\newcommand{\modelname}{SIMBA}

\DeclareMathOperator*{\concat}{\scalerel*{\Vert}{\sum}}

\ccsdesc[500]{Computing methodologies~Neural networks}
\ccsdesc[500]{Computing methodologies~Learning latent representations}

\title{Graph Size-imbalanced Learning with Energy-guided         
Structural Smoothing
}

\begin{document}

\author{Jiawen Qin}
\affiliation{%
  \institution{Beihang Univerisity}
  \city{Beijing}
  \country{China}}
\email{qinjw@buaa.edu.cn}

\author{Pengfeng Huang}
\affiliation{%
  \institution{Beihang Univerisity}
  \city{Beijing}
  \country{China}}
\email{huangpf@act.buaa.edu.cn}

\author{Qingyun Sun}
\affiliation{%
  \institution{Beihang Univerisity}
  \city{Beijing}
  \country{China}}
\email{sunqy@buaa.edu.cn}

\author{Cheng Ji}
\affiliation{%
  \institution{Beihang Univerisity}
  \city{Beijing}
  \country{China}}
\email{jicheng@act.buaa.edu.cn}

\author{Xingcheng Fu}
\affiliation{%
  \institution{Guangxi Normal University}
  \city{Guilin}
  \country{China}}
\email{fuxc@gxnu.edu.cn}

\author{Jianxin Li}
\affiliation{%
  \institution{Beihang Univerisity}
  \city{Beijing}
  \country{China}}
\email{lijx@buaa.edu.cn}


\renewcommand{\shortauthors}{Jiawen Qin, et al.}

\begin{abstract}
Graph is a prevalent data structure employed to represent the relationships between entities, frequently serving as a tool to depict and simulate numerous systems, such as molecules and social networks.
However, real-world graphs usually suffer from the size-imbalanced problem in the multi-graph classification, i.e., a long-tailed distribution with respect to the number of nodes.
Recent studies find that off-the-shelf Graph Neural Networks (GNNs) would compromise model performance under the long-tailed settings.
We investigate this phenomenon and discover that the long-tailed graph distribution greatly exacerbates the discrepancies in structural features.
To alleviate this problem, we propose a novel energy-based size-imbalanced learning framework named \textbf{SIMBA}, which smooths the features between head and tail graphs and re-weights them based on the energy propagation.
Specifically, we construct a higher-level graph abstraction named \textit{Graphs-to-Graph} according to the correlations between graphs to link independent graphs and smooths the structural discrepancies.
We further devise an energy-based message-passing belief propagation method for re-weighting lower compatible graphs in the training process and further smooth local feature discrepancies.
Extensive experimental results over five public size-imbalanced datasets demonstrate the superior effectiveness of the model for size-imbalanced graph classification tasks.
\end{abstract}

\keywords{Graph representation learning, graph neural networks, imbalance learning, graph classification}

\maketitle

\section{INTRODUCTION}
Graphs, or networks, refer to interconnected structures that are commonly used to model interaction between entities such as social networks~\cite{girvan2002community} or molecules~\cite{cai2018comprehensive}.
Graph representation learning arises as a prevalent tool to project graphs into low dimensional vector representations due to the emergence of Graph Neural Networks (GNNs)~\cite{kipf2017semi, petar2018graph, gilmer2017neural,tao2022exploring, zhou2023greto} associated with its effectiveness in expressing graph structures.
GNNs conduct neighborhood aggregation operations to recursively pass and aggregate messages along edges to acquire the node representations in an end-to-end manner, thus encoding both graph structure and node context information. 
These previous approaches thereby establish the basis for acquiring graph-level representation learning~\cite{zhang2018end, ma2019graph} and numerous GNN variants have been proposed by configuring different propagation and pooling schemes, enabling several downstream tasks such as multi-graph classification~\cite{ying2018hierarchical, lee2019self}.

\begin{figure}[!t]
\centering
\subfigure[{Long-tailed graph distribution.}]{
\includegraphics[width=0.48\linewidth]{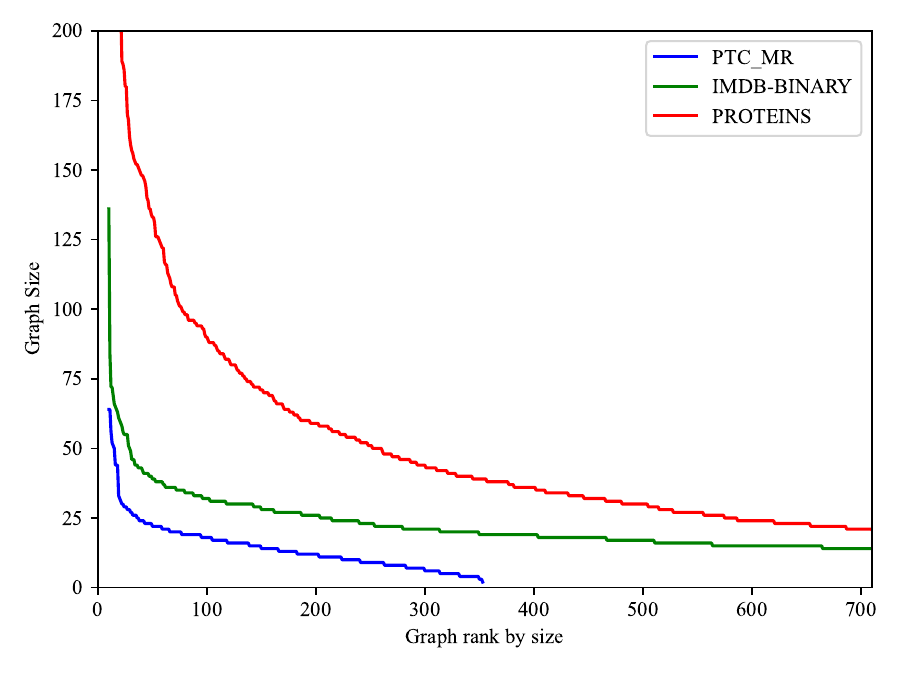}
\label{fig:graph size}
}%
\subfigure[{Structural discrepancies in different graph distribution.}]{
\includegraphics[width=0.48\linewidth]{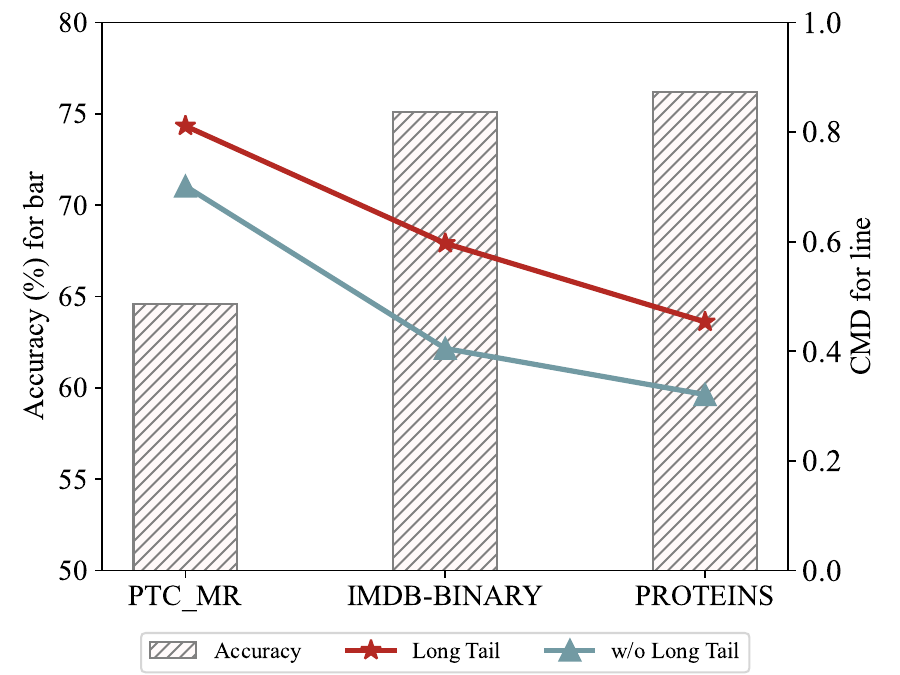}
\label{fig:structural discrepancies}
}%
\centering
\caption{(a) The graph sizes in three benchmark datasets follow the power-law-like distribution. (b) We apply GIN for acquiring structural features of graphs and use the Central Moment Discrepancy (CMD)~\cite{zellinger2019robust} to evaluate the structural discrepancies under the long-tail and relative size-balanced graph distributions.
A higher CMD score indicates a larger variation in feature distribution.}

\vspace{-0.3cm}
\label{fig:introduction}
\Description{}
\end{figure}

\textbf{Problem.}
In many real-world graph datasets, the graph sizes follow a power-law-like distribution, with a large portion of the graphs having relatively small sizes, as shown in Figure~\ref{fig:graph size}.
A few head graphs occupy a handful of large sizes, while a significant fraction of tail graphs have small sizes, illustrating the \textbf{size-imbalanced} problem.
Although GNNs are designed to be able to work on graphs of any size~\cite{buffelli2022sizeshiftreg}, the graphs with larger size usually achieve higher accuracy, while the performance gradually drops as the graph size decreases~\cite{liu2022size}.
Their large disparities performed on graphs with varying sizes  severely impact on the generalization of models.
As shown in Figure~\ref{fig:structural discrepancies}, applying GIN~\cite{xu2018powerful} for graph classification, the long-tailed graph distribution exacerbates the discrepancies in structural features across graphs.
The larger variation in structural distribution gives rise to the inferior generalization to the model.

\textbf{Research Background.}
Although imbalance learning on graphs has recently attracted many research interests~\cite{mandal2022metalearning,iglbench2024}, most of them focus on the node-level imbalance classification issue, including class imbalance~\cite{zhao2021graphsmote, park2021graphens, song2022tam} and structure imbalance~\cite{gao2021topology, sun2022position, fu2023hyperbolic}.
The size-imbalanced learning for graph-level classification is still under-explored.
The only existing work, SOLT-GNN~\cite{liu2022size}, assumes the structures in the head graphs are sufficient for discrimination and enhances latent representations of tail graphs by transferring structural information from head to tail graphs.
However, the strong assumption leads to poor generalization and unsatisfied overall performance for size-imbalanced learning.
While benefiting tail graphs, SOLT-GNN inadvertently creates an artificial out-of-distribution scenario during training~\cite{buffelli2022sizeshiftreg}, where models primarily observe tail graphs.
Because the size range of head and tail graphs differs substantially among graph datasets, optimizing the performance of predefined tail graphs by transferring the knowledge of certain head graphs does not essentially smooth the structural discrepancies across graphs.
Moreover, after the multiple layers of propagation and pooling in the GNN architectures, the learned embeddings are too coarse to reserve the structural information of graphs with long-tailed size distribution, which further restricts the model's expressiveness.
To enhance the generalization capacity of GNNs in long-tailed graph distribution, learning discriminative graph embeddings and smoothing the structural features are required to achieve a superior representation.

\textbf{Major Challenges.} 
Unlike node classification utilizing the link structure of embedded nodes, graphs are individual instances that are isolated from each other in multi-graphs classification task, and thus directly aggregating and smoothing features across graphs
remains challenging.
This non-trivial solution presents two main challenges for size-imbalanced learning on graph-level classification. 
First, \textit{how to establish connections between graphs (including both the head and tail ones) according to their discriminative representations?}
Second, \textit{how to propagate and smooth structural features among graphs in the training process?
}

\textbf{Present Work.} 
To tackle these challenges, in this paper, we propose a novel \underline{\textbf{S}}ize-\underline{\textbf{IMBA}}lanced learning framework, \textbf{\modelname}, to narrow the gap between head and tail graphs.
In particular, \modelname~ attempts to smooth the structural discrepancies in head and tail graphs, which can further promote the generalization capacity of the model and boost the overall performance.
To deal with the first challenge, we propose a size-invariable graph embedding technique to capture discriminative information of the graph, especially from node- and layer-levels. 
We further construct a higher-level graph abstraction named graphs-to-graph according to the correlations of graphs to link independent graphs and perform message passing operations to aggregate the information from neighboring graphs.
Aggregating representations of graphs simulates feature propagation and aggregation in node classification tasks, which smooths the features across graphs.
To address the second challenge, we devise an energy-based graph re-weighting mechanism to influence the loss function by measuring the local compatibility of each graph in graph abstraction and assigning relatively higher costs to instances with minor energy scores.
The energy-based belief of each graph over constructed graph abstraction recursively propagates among neighboring graphs and aggregates energy scores.
The propagation scheme pushes the energy towards the majority of neighboring graphs, which contributes to amplifying the energy gap between graphs that play a negative or positive role in the training process.
Overall, the main contributions are summarized as follows: 
\begin{itemize}[leftmargin=*]
\item We propose a novel framework \modelname~ to promote the generalization capacity of the model for size-imbalanced graph classification, which adapts to different levels of size-imbalanced scenarios. 
\item To mitigate structural discrepancies in the size-imbalanced graph distribution, \modelname~ smooths the features among neighboring graphs on constructed graphs-to-graph unit and adjusts the influence of each graph in the local region by 
performing the energy propagation based on the energy-based belief measure.
\item Experimental results on real-world size-imbalanced datasets demonstrate \modelname~ can outperform the state-of-the-art baselines and and enhance the generalization to head and tail graphs.
\end{itemize}

\section{RELATED WORKS}
In this section, we provide an analysis of the related works on graph representation learning and imbalance learning on graphs.

\textbf{Graph Representation Learning.}
Graph representation learning~\cite{bryan2014DeepWalk, grover2016node2vec} has demonstrated its power for graph analytics. 
Recently, more focus has been shifted to graph neural networks (GNNs)~\cite{kipf2017semi, hamilton2017inductive, petar2018graph, fu2023adaptive}, which recursively aggregate neighborhood information and pass messages along edges to learn node representations. 
A typical GNN architecture for graph classification begins with message-passing layers that extract node representations followed by pooling layers, which gradually condense node representations into graph representations and then predict graph labels. 
Graph pooling operations~\cite{xu2018powerful} boost the investigation of graph-level representation and can be roughly divided into flat pooling and hierarchical pooling~\cite{liu2022graph}. 
Flat pooling methods~\cite{zhang2018end, wu2019net, wang2020second, baek2021accurate} are also known as readout operation, which summarizes all node representations in a pooling layer, such as sum-pooling and mean-pooling. 
Hierarchical pooling approaches~\cite{yuan2020structpool, ranjan2020asap} try to gradually preserve the graph's hierarchical structure information by recursively coarsening the graph into a smaller size.
Several approaches are presented to select the most important $k$ nodes from the original graph to generate a new one, such as TopKPool~\cite{gao2019graph}, SAGPool~\cite{lee2019self} and TAPool~\cite{gao2021topology}. 
Another design groups nodes into clusters and generates new nodes for the coarsened graph, including DiffPool~\cite{ying2018hierarchical}, MinCutPool~\cite{bianchi2020spectral} and SEP~\cite{wu2022structural}.
However, the above graph-level classification methods still suffer from the size-imbalanced issue, which is still under-explored.

\textbf{Imbalance learning on graphs.}
Imbalance learning~\cite{he2009learning} has attracted extensive research attention in the field of graph data, which is complicated due to various graph structures.
In addition to the imbalanced number of labeled instances, imbalance learning on graphs can spring from differences in structural abundance across groups.
Most existing works are dedicated to node-level imbalance problem~\cite{shi2020multi, li2023graphsha}, which can be divided into two main streams, re-weighting and re-sampling.
Re-sampling strategies~\cite{zhao2021graphsmote, qu2021imgagn, park2022graphens, yun2022lte4g} balance the data distributions by up-sampling minority classes, under-sampling majority classes or generating new samples for minority classes. 
GraphSMOTE~\cite{zhao2021graphsmote} attempts to generate edges by pretraining an edge generator for isolated synthetic nodes generated from SMOTE~\cite{chawla2002smote}. 
GraphSHA~\cite{li2023graphsha} synthesizes harder minor samples and avert invading neighbor subspaces when enlarging the minor subspaces.
Re-weighting strategies~\cite{song2022tam, xia2022cengcn, sun2022position, fu2023hyperbolic} attempt to modify the loss function by raising the weights of minority classes, depressing the weights of majority classes or expanding the margins on minority classes.
HyperIMBA~\cite{fu2023hyperbolic} re-weights the labeled nodes based on their 
relative positions to class boundaries to alleviate the hierarchy-imbalanced issue.

However, all of these recent powerful deep learning works are proposed for imbalanced issues in node classification. 
The graph-level imbalanced issue remains largely unexplored. 
~\cite{wang2022imbalanced,imgkb2023} focus on class-imbalanced issue.
G$^2$GNN~\cite{wang2022imbalanced} derives extra supervision globally from neighboring graphs with similar topology and locally from stochastic augmentations of graphs.
However, these constructed neighbors ignore the potential problem of size imbalance and cannot smooth the structural discrepancies in the training process.
SOLT-GNN~\cite{liu2022size} identifies long-tailed size-imbalanced graph classification problem and focuses on promoting the performance on tail graphs by transferring the structural knowledge of head graphs to tail graphs, where the performance is naturally limited by the predefined ratio of head graphs.
In this work, we aim to find a solution to alleviate the impact of the size-imbalanced issue on both the head and tail graphs.

\section{PRELIMINARY}
In this section, we provide the necessary background information and notations about size-imbalance learning for graph classification.

\subsection{Problem Formulation}
\label{definition}

\textbf{Graph.}
A graph is denoted as $G=\{\mathcal{V},\mathcal{E}, \mathbf{X}\}$, where $\mathcal{V}$ is the node set and $|\mathcal{V}|=n$, $\mathcal{E}$  denotes the edge set, and $\mathbf{X}\in\mathbb{R}^{n\times d}$ is the feature matrix  for all nodes with $d-$dimension feature vector.
Let $\mathbf{A}\in\mathbb{R}^{n\times n}$ denote the adjacency matrix, where $\mathbf{A}\in\{0,1\}^{n\times n}$ for unweighted graphs. 
The element $\mathbf{A}_{ij}$ is equal to 1 if there exists an edge between node $v_{i}$ and node $v_{j}$, and zero otherwise.

Given a set of $N$ graphs $\mathcal{G}=\{G_1,G_2,\ldots,G_N\}$ and their labels $\mathbf{Y}\in\mathbb{R}^{N\times C}$, where each graph is defined as $G_i=\{\mathcal{V}_i,\mathcal{E}_i, \mathbf{X}_i\}$, $C$ is the number of classes.
The graph sizes $|G_i|$, $i.e.$, the number of graph nodes, usually follow a power-law-like distribution where a significant fraction of graphs have small sizes. 
We sort all graphs in descending order by their sizes.
The first $T$ graphs are called head graphs, $i.e.$, $\mathcal{G}_{\mathrm{head}}=\{G_1,G_2,\ldots,G_T\}$, and the remaining graphs are called tail graphs, $i.e.$, $\mathcal{G}_{\mathrm{tail}}=\{G_{T+1},G_{T+2},\ldots,G_N\}$.
To measure the extent of size imbalance in different datasets, we propose the size-imbalanced ratio (SIR) as  $SIR = \frac{\frac1{|\mathcal{G}_{\mathrm{head}}|}\sum|{G}_i|,{G}_i\in\mathcal{G}_{\mathrm{head}}}{\frac1{|\mathcal{G}_{\mathrm{tail}}|}\sum|{G}_j|,{G}_j\in\mathcal{G}_{\mathrm{tail}}}$.

\textbf{The problem.}
Given a size-imbalanced multi-graphs dataset $\mathcal{G}$, we aim to learn a powerful classifier $f_\psi~:~\mathcal{G}~\to~\mathbb{R}^C$ for predicting the label of each $G_{i}\in \mathcal{G}$ according to its representation $\mathbf{H}_{G_i}$ that works well for both head and tail graphs.

\subsection{Graph Neural Networks}
Graph neural networks (GNNs)~\cite{cai2018comprehensive} typically conduct message passing and aggregation operations recursively from the neighboring nodes along the edges to form the node representations in an end-to-end manner. 
A generic GNN layer stack can be denoted as:
\begin{align}
\mathbf{h}_v^{(l)}=\operatorname{COM}^{(l)}\left(\mathbf{h}_v^{(l-1)},\operatorname{AGG}^{(l)}\{\mathbf{h}_u^{(l-1)}|u\in\mathcal{N}(v)\}\right), 
\label{Eq:GNN}
\end{align}
where $\mathcal{N}(v)$ is the neighbors set of node $v$, $\operatorname{AGG}^{(l)}$ is a neighborhood aggregation function in layer $l$, and $\operatorname{COM}^{(l)}$ is a function that combines the own features with the neighboring node features in layer $l$.
In the input layer, the node's initial representation is denoted as $\mathbf{h}_{v}^{0}=\mathbf{x}_v$. 
The output of a $L$-layer convolutions for node $v$ is $\mathbf{h}_{v}^{(L)}$.
For graph classification tasks, a $\operatorname{READOUT}$ layer typically transforms the feature vectors from the last layer to generate the final output $\sum_{v\in 
\mathcal{V}}\mathbf{h}_{v}^{(L)}$.

\subsection{Energy-based Model}
Energy-based models (EBMs)~\cite{ranzato2006efficient, xie2016theory} have been introduced to measure the compatibility between observed variables and variables to be predicted using an energy function.
The energy function $E(g,y):\mathbb{R}^d\to\mathbb{R}$ maps each $d$-dimensional input instance $g$ with an arbitrarily specific class label $y$ to a scalar value ($i.e.$, energy).
Recent works~\cite{grathwohl2020your, wu2023energy} establish an equivalence between a neural classifier and an energy-based one.
Given a classifier as $\begin{aligned}f(g):\mathbb{R}^d\to\mathbb{R}^C\end{aligned}$ which is mapping an input instance with $d$-dimensional feature to a $C$-dimension logits. 
A softmax function over the logits gives a predictive categorical distribution: $p(y|{g})=\frac{e^{f({g})_{[y]}}}{\sum_{c=1}^Ce^{f({g})_{[c]}}}$, where $f({g})_{[c]}$ queries the $c$-th entry of the output. 
The equivalence between an EBM and a neural classifier is formed by setting the energy as the predicted logit value $E({g},y)=-f({g})_{[y]}$.
The energy function $E({g})$ for any given instance can be: 
\begin{align}
E({g})=-\log\sum_{y^{\prime}}e^{-E({g},y^{\prime})}.
\label{Eq:energy}
\end{align}
\hspace{-1.5em}
\begin{figure*}[htb]
\centering
\includegraphics[width=1\textwidth]{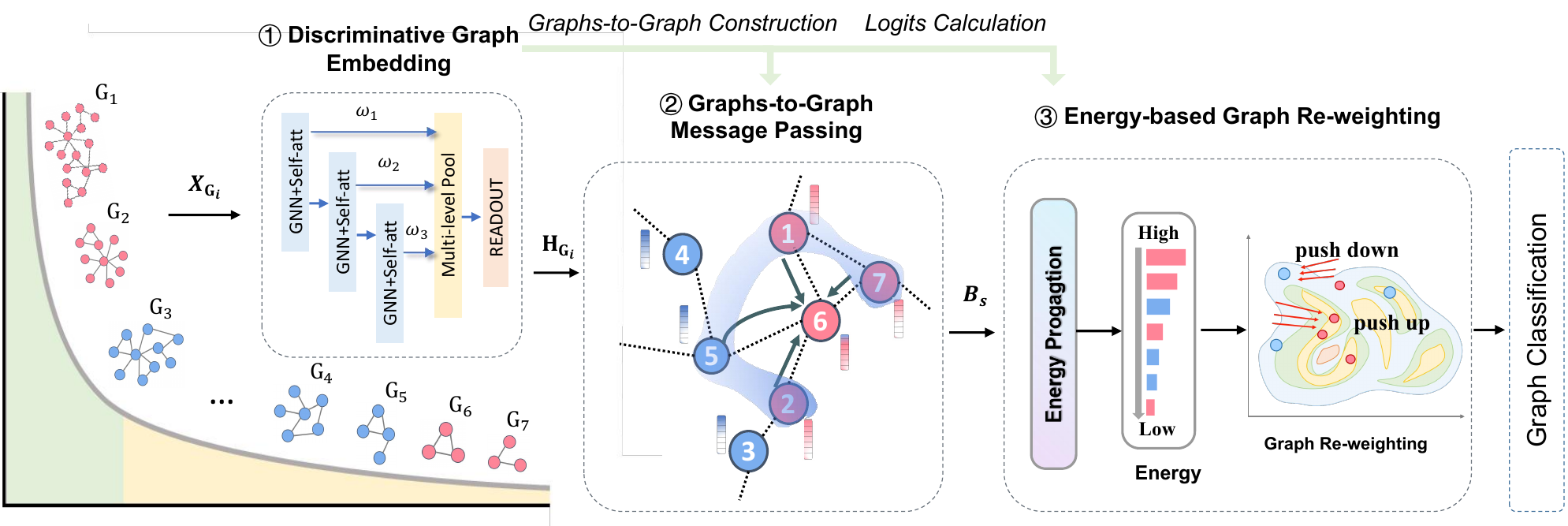}
\vspace{-1.2em}
\caption{\textbf{An illustration of \modelname~ architecture.} 
(1) \modelname~ derives the graph's representations through a discriminative graph encoder.
(2) \modelname~ establishes a higher-level graph abstraction (graphs-to-graph) to acquire extra supervision structural information from the top-$k$ nearest graphs in the latent space.
(3) \modelname~ propagates energy belief of each instance along the edges over constructed graphs-to-graph and re-weight it according to the ordered energy score list of all instances. }
\vspace{-1em}
\label{fig:Architecture}
\Description{}
\end{figure*}

\section{THE PROPOSED FRAMEWORK}
In this section, we present the proposed framework \modelname~, which is composed of two modules. 
First, in order to connect independent graphs, \modelname~ establishes a higher-level graph abstraction named graphs-to-graph, enabling two-level propagation.
This is achieved by deriving graph representations through a size-invariable graph encoder in Section \ref{pooling} and subsequently aggregating neighboring graph representations via  graphs-to-graph propagation in Section \ref{GoG}.
Second, to further smooth the structural features in local regions, \modelname~ presents
an energy-based belief propagation, which propagates beliefs along edges over constructed graphs-to-graph unit, and re-weights each graph according to the estimated energy score to make more compatible graphs among neighboring graphs play more active roles during the model learning progress in Section \ref{reweight}.
The overall framework of \modelname~ is illustrated in Figure \ref{fig:Architecture}. 
In the following part, we will elaborate on the details.

\subsection{Discriminative Graph Embedding}
\label{pooling}
Due to the long-tailed distribution of graph sizes, the expressiveness of the model is limited by the fixed embedding dimension, leading to information loss. 
To reserve structural information of long-tailed graphs, we present a discriminative graph embedding technique that facilitates the derivation of a fixed-length embedding vector for each variable-sized graph and acquires the structural difference at multiple levels. 
We first use a GNN layer to transform the initial node’s features and then utilize a self-attentive mechanism~\cite{lin2017structured} to capture discriminative information and transform the size-variable nodes into a fixed-length embedding vector in each convolutional layer. 
Finally, layer-wise pooling operations derive graph representations by aggregating information with multiple-scale topology. 

\textbf{Step1: Size-invariable Node Embedding.}
We employ a GNN model as the encoder to learn graph representation for each graph $G_{i}\in \mathcal{G}$ and our framework is agnostic to the base GNN models and flexible to most message passing neural networks.

Note that, the node embeddings at $l$-th convolutional layer denote as  $\mathbf{h}_{G_i}^{(l)}\in\mathbb{R}^{{n_i}\times{r}}$, which is the input of $l$-th pooling layer.
The embeddings are induced by the number of nodes $n_i$, $i.e.$, the size of the graph $G_i$.
We then use the self-attentive mechanism to learn discriminative information of nodes and embed it to a fix-length vector, which is size invariant:
\begin{align}
    \mathbf{M}^{(l)}=\operatorname{softmax}(\mathbf{W}_{2}^{(l)}\operatorname{tanh}(\mathbf{W}_{1}^{(l)}(\mathbf{h}_{G_i}^{(l)})^\top)),
\label{eq:invariant}
\end{align}
where $\mathbf{W}^{(l)}_1\in\mathbb{R}^{{q}\times{r}}$ linearly transforms the node representations to a $q$-dimensional space, and followed by a nonlinear function $\operatorname{tanh}$.
$\mathbf{W}^{(l)}_{2}\in\mathbb{R}^{{1}\times{q}}$ is utilize to learn the differences between nodes based on $l$-hop information centered at each node.
The input of $l$-th pooling layer, $i.e.$, the output of $l$-th concolutional layer is computed by multiplying $\mathbf{M}^{(l)}\in\mathbb{R}^{{1}\times{n}}$ with $\mathbf{h}_{G_i}^{(l)}\in\mathbb{R}^{{n}\times{h}}$, which is size invariant due to it does not depend on the graph sizes.

\textbf{Step-2: Multi-level Pooling.}
Instead of directly computing the graph representation from the final nodes representations $\mathbf{h}_{G_i}^{(L)}$, we employ a multi-level pooling architecture, which computes layer-wise graph representations and captures the structural difference between multiple levels.
The graph-level embedding is:
\begin{align}  \mathbf{H}_{G_i}=\operatorname{MLP}_{\phi}\left(\displaystyle{\concat_{l=1}^{L}}\omega^{(l)}\mathbf{M}^{(l)}\mathbf{h}_{G_i}^{(l)}\right),
\label{eq:pool}
\end{align}
where $\concat$ is a concatenation operation, $L$ is the number of convolutional layers, $\omega^{(l)}$ is a trainable weight vector, and $\operatorname{MLP}_{\phi}$ denotes a two-layer perceptron applied on the concatenated results.

\subsection{Graphs-to-Graph Message Passing}
\label{GoG}
Graph representations acquired by solely forwarding each graph through GNN encoders cannot generalize to the long-tailed graph classification.
The large structural discrepancies between head and tail graphs compromise the model’s performance.
We construct a  graphs-to-graph unit to pull graphs with similar embeddings closer together and  explicitly integrate various types of inter-dependencies among the original graphs.
Graphs with the same category, including head and tail graphs, exchange information and smooth the structural discrepancies, making it easier for the classifier to discriminate instances.
The intuition is that transferring information of graphs in graphs-to-graph unit simulates feature propagation and aggregation in node classification tasks, which
smooths the features cross graphs. 
Performing message passing and aggregating operations from the neighboring graphs supplements the structural information of both head and tail graphs, narrowing the gap between them.

\textbf{Graphs-to-Graph Unit.}
Given a set of graphs $\mathcal{G}$, we aim to construct a higher-level graph abstraction (graphs-to-graph) where each graph $G_{i}\in \mathcal{G}$ is denoted as a node and two graphs are connected by an edge if they are similar.
After obtaining the embeddings of graphs in Section~\ref{pooling}, we utilize a metric function that accounts for graph embedding information similarities to measure the likelihood of edge existence between pairs of graphs. 
In this work, we leverage the widely used cosine similarity function as follows:
\begin{align}
S_{ij}=\frac{\mathbf{H}_{G_i}^\top\mathbf{H}_{G_j}}{\|\mathbf{H}_{G_i}\|\|\mathbf{H}_{G_j}\|}.
\label{Eq:similar}
\end{align}
Then we choose $k$ nearest neighbors for each graph following the above cosine similarity and construct the $k$-NN graph by linking each graph $G_{i}$ with its top-$k$ similar graphs.
We denote the adjacency matrix of the constructed graphs-to-graph as $\mathbf{B}_s$ and its corresponding degree matrix as $\mathbf{D}_s$.

\textbf{Graphs-to-Graph Propagation.}
After we obtain the higher-level graph abstraction, the message propagation for each instance on graphs-to-graph at the $l$-th layer can be formulated as:
\begin{align}
\mathbf{H}^{(l)}_{{G}_i}=\tilde{\mathbf{D}}_{s}^{-1}\tilde{\mathbf{B}}_{s}\mathbf{H}^{(l-1)}_{{G}_{i}},
\label{Eq:propagate}
\end{align}
where $\tilde{\mathbf{B}}_{s}$ is the adjacency matrix with added self-loops of the constructed new graph abstraction, $\mathbf{H}^{(0)}_{{G}_{i}}$ represents the graph embedding previously derived from the pooling operation in Section \ref{pooling}.
After $l$ layers message passing, the representation of an individual graph aggregates information from neighboring graphs up to $l$-hops away, which naturally acquire extra structural information from their similar graphs and smooth the features.

\subsection{Energy-based Graph Re-weighting}
\label{reweight}
Even though each graph globally derives extra structural  information from its neighboring graphs, under-represented tail graphs suffering from scarce structure information propagate low belief along the edges, exacerbating the impact derived from long-tailed graph distribution in the local region.
We design an energy-based graph re-weighting mechanism to guide the model towards achieving a harmonious low-energy state and improve its generalization ability. 
It allows for incompatible tail graphs among neighboring graphs to play a non-activating roles during model training, further smooth the structural discrepancies between graphs.

\begin{algorithm}[t]
    \LinesNumbered
    \caption{\modelname} 
    \label{Algorithm}
    \KwIn{Graph with labels $\{\mathcal{G} = (\mathcal{V},\mathcal{E}, \mathbf{X}, \mathbf{A}), Y\}$; Number of training epochs $n_e$; Convolutional layers $L$.}
    \KwOut{Predicted graph label $\hat{Y}$.}
    \For{$t = 1,2,\cdots,n_e$}{
        \tcp{Graph Encoding}
        \For{$l = 1,2,\cdots,L$}{
        Calculate the set of node representation $h$ at $l$-th layer by a GNN layer; \\
        Encode layer-wise graph representation into a fix-length embedding vector by multiplying $\mathbf{M}^{(l)}$ (Eq.(\ref{eq:invariant})) with  $\mathbf{h}_{G_i}^{(l)}$;     \\
        }
        Readout the graph representation $\mathbf{H}_{G_i}$ by Eq.(\ref{eq:pool}); \\
        \tcp{Graphs-to-Graph Message Passing}
        Find top-$k$ similar graphs for each $G_{i}\in \mathcal{G}$ by Eq.(\ref{Eq:similar}) \\
        Apply the Graphs-to-Graph propagation by Eq.(\ref{Eq:propagate});
        \\
        \tcp{Energy-based Graph Re-weighting}
        Calculate the energy score for  each $G_{i}\in \mathcal{G}$ over constrcted graphs-to-graph by Eq.(\ref{Eq:free energy});\\
        Perform the energy propagation along the edges and update energy scores for graphs by Eq.(\ref{eq:energy propa});  \\
        Calculate the training weight for each $G_{i}\in \mathcal{G}$ by Eq.(\ref{eq:reweight});\\
        \tcp{Optimization}
        Calculate the overall loss $\mathcal{L}$ by Eq.(\ref{eq:loss}); \\
        Update model parameters to minimize $\mathcal{L}$; \\
    }        
\end{algorithm}

\textbf{Energy-based Belief Propagation.}
Recently, energy-based models (EBMs)~\cite{grathwohl2020your, wu2023energy} have been introduced to measure the compatibility between observed variables and variables to
be predicted using an energy function.
With $l$ layers message passing in graphs-to-graph unit, each graph $G_{i}\in \mathcal{G}$ derives a vector $\mathbf{u}_{i}^{(l)}$ as logits, formulated as:
\begin{equation}
    f_\psi({G}_i,\mathcal{R}^{(l)}_{{G}_i})=\mathbf{u}_i^{(l)},
\end{equation}
where $\psi$ means the trainable parameters of the classifier $f$.
According to the graphs-to-graph propagation, the predicted logits, for instance, $G_i$ are formulated as the $l$-order neighboring graphs centered at instance $G_i$, denoted as $\mathcal{R}^{(l)}_{{G}_i}$.
The logits are utilized to obtain a categorical distribution for classification:
\begin{align}
p\left(y\mid{G}_i,\mathcal{R}^{(l)}_{{G}_i}\right)=\frac{e^{f_\psi({G}_i,\mathcal{R}^{(l)}_{{G}_i})_{[y]}}}{\sum_{c=1}^Ce^{f_\psi({G}_i,\mathcal{R}^{(l)}_{{G}_i})_{[c]}}}.
\label{Eq:softmax}
\end{align}
By connecting it to the EBM model that defines the relation between energy function and probability density, the energy form is denoted as $E({G}_i,\mathcal{R}^{(l)}_{{G}_i},y;f_\psi)=-f_\psi({G}_i,\mathcal{R}^{(l)}_{{G}_i})_{[y]}$.
And the free energy function $E({G}_i,\mathcal{R}^{(l)}_{{G}_i};f_\psi)$ that marginalizes $y$ can be represented as the denominator in Eq.\eqref{Eq:softmax}:
\begin{align}
E\left({G}_i,\mathcal{R}^{(l)}_{{G}_i}; f_{\psi}\right)=-\log\sum_{c=1}^{C}e^{f_{\psi}({G}_i,\mathcal{R}^{(l)}_{{G}_i})_{[c]}}.
\label{Eq:free energy}
\end{align}
Inspired by label propagation on the graph~\cite{zhu2003semi}, we use the estimated energy scores based on energy-based belief propagation to measure the compatibility of graphs among neighboring graphs.
The rationale behind the design of energy belief propagation is to accommodate the physical mechanism in data generation with instance-wise interactions. 
The input graphs in graphs-to-graph unit could reflect the certain proximity on the manifold.
The energy-based belief of each graph over constructed graph abstraction recursively propagates among neighboring graphs.

Specifically, we define $\mathbf{E}^{(0)}$ as a vector of initial energy scores for graphs in graphs-to-graph unit, and the energy propagation along the edges can be formulated as:
\begin{align}
    \mathbf{E}^{(0)}&=[E({G}_i,\mathcal{R}^{(l)}_{{G}_i};f_\psi)]_{{G_i}\in\mathcal{G}}, \\
    \mathbf{E}^{(t)}&=\lambda\mathbf{E}^{(t-1)}+(1-\lambda)\tilde{\mathbf{D}}_{s}^{-1}\tilde{\mathbf{B}}_{s}\mathbf{E}^{(t-1)}, \\
    \mathbf{E}^{(t)}&=\left[E_{G_i}^{(t)}\right]_{{G_i}\in\mathcal{G}},
\label{eq:energy propa}
\end{align}
where $\lambda\in(0,1)$ is a hyper-parameter governing the concentration on the energy of the graph itself and other connected graphs, $\mathbf{E}^{(t)}$ is the final estimated energy score for each graph in constructed graphs-to-graph after $t$-step message passing.

The propagation scheme would push the energy towards the compatible graphs with similar structural distribution, which can help amplify the energy gap the incompatible graphs with high-energy scores among neighboring graphs.
This finding implies that the energy model can be enhanced through the propagation scheme during inference, without incurring any additional costs for training.
We will verify its effectiveness through empirical comparison in Section \ref{sec:Experiment}.

\textbf{Instance-wise Graph Re-weighting.}
To mitigate the impact of an incompatible graph on its neighbors in graphs-to-graph unit, which does not align with the structural distribution in local regions, we introduce a cosine annealing mechanism to re-weight each instance in graphs-to-graph according to the estimated energy score propagating along edges. 
The training weights can be formulated as: 
\begin{align}
    \delta_{G_i}= \epsilon_\mathrm{min}+\frac12\left( \epsilon_\mathrm{max}- \epsilon_\mathrm{min}\right)\left(1+\cos\left(\frac{\mathcal{Q}\left(E_{G_i}^{(t)}\right)}{|\mathcal{G}|}\pi\right)\right),
\label{eq:reweight}
\end{align}
where $\delta_{G_i}$ is the modified training weight for the graph $G_i$. 
$\epsilon_\mathrm{min}$, $\epsilon_\mathrm{max}$ are the hyper-parameters stating the lower bound and upper bound of the weight correction factor.
$\mathcal{Q}(E_{G_i}^{(t)})$ denotes the ordered list of $E_{G_i}^{(t)}$ from smallest to largest.
The training loss for the size-imbalance graph classification task is computed by minimizing the negative log-likelihood of labeled training data and formulated as:
\begin{align}
    \mathcal{L}=\sum_{G_i\in\mathcal{G}}\delta_{G_i}\left(-f_\psi\left({G}_i,\mathcal{R}^{(l)}_{{G}_i}\right)_{[y_i]}+\log\sum_{c=1}^Ce^{h_\psi({G}_i,\mathcal{R}^{(l)}_{{G}_i})_{[c]}}\right).
\label{eq:loss}
\end{align}
The overall process of the proposed framework \modelname~ is shown in Algorithm~\ref{Algorithm}.

\section{EXPERIMENT}
\label{sec:Experiment}

\subsection{Experimental Setups}
\subsubsection{Datasets.}

We utilize five widely adopted real-world datasets, comprising four from bioinformatics  and one from social networks. 
The bioinformatics datasets include PTC-MR, D\&D, PROTEINS, and FRANKENSTEIN, while the social network dataset is REDDIT-B.
Statistics of these datasets can be found in Table \ref{dataset_description}.
\begin{table}[!ht]
\small
\vspace{-1em}
\caption{Statistics of datasets.}
\vspace{-1em}
\centering
\scalebox{0.96}{
\begin{tabular}{l|rrrrrr}
\toprule
{Datasets}  & {\#Graphs} & {Avg. $|\mathcal{V}|$} &{Avg. $|\mathcal{E}|$} & {\#Cla.} & {\#Attr.}  & $T$\\ 
\midrule
{PTC-MR}     & 344  & 14.29 & 14.69  & 2     & 18 & 67 \\
{FRANK.}  & 4,337 & 16.90 & 17.88  & 2    & 5  & 757  \\
{D\&D}    & 1,178  & 284.32 & 715.66  & 2   & 89 & 234 \\
{PROTEINS}     & 1,113  & 39.06 & 72.82  & 2     & 3 & 222 \\
{REDDIT-B}    & 2,000  & 429.63 & 497.75  & 2   & 566 & 400 \\
\bottomrule
\end{tabular}
}
\vspace{-1em}
\label{dataset_description}
\end{table}

\begin{table*}[htbp]
\vspace{-1em}
\caption{Size-imbalanced graph classification accuracy and Macro-F1 score (
\%) with the standard errors on five datasets.
(Result: \textbf{Bold}: best; \underline{Underline}: runner-up.)}
\vspace{-1em}
\centering
\resizebox{2.1\columnwidth}{!}{
\begin{tabular}{c|cc|cc|cc|cc|cc}
\toprule
\multirow{2}{*}{\textbf{Method}} & \multicolumn{2}{c|}{\textbf{PTC-MR}}    & \multicolumn{2}{c|}{\textbf{FRANKENSTEIN}}    & \multicolumn{2}{c|}{\textbf{D\&D}}  & \multicolumn{2}{c|}{\textbf{PROTEINS}}   & \multicolumn{2}{c}{\textbf{REDDIT-B}}   \\ 
                        & Acc.                & M-F1                     & Acc.                & M-F1        & Acc.                & M-F1        & Acc.                & M-F1        & Acc.                & M-F1\\ 
\midrule
SP    & 56.00\scalebox{0.75}{±1.40} & 47.29\scalebox{0.75}{±3.56}         & 59.10\scalebox{0.75}{±1.85}   & 53.01\scalebox{0.75}{±1.93}      &\underline{75.30\scalebox{0.75}{±1.61}}    & \underline{72.41\scalebox{0.75}{±1.78}}   &73.66\scalebox{0.75}{±1.86}   &71.13\scalebox{0.75}{±1.99} &75.05\scalebox{0.75}{±1.76} &74.78\scalebox{0.75}{±1.63}\\ 
GK  & \underline{58.16\scalebox{0.75}{±4.90}} & 46.75\scalebox{0.75}{±6.02} & 57.83\scalebox{0.75}{±1.86}   & 50.36\scalebox{0.75}{±2.60}       & 72.57\scalebox{0.75}{±1.13}   & 69.27\scalebox{0.75}{±4.78}  &71.25\scalebox{0.75}{±1.46}  & 66.32\scalebox{0.75}{±2.42}  &71.75\scalebox{0.75}{±2.17}  &70.25\scalebox{0.75}{±2.70}\\ 
\midrule
GCN    & 54.29\scalebox{0.75}{±4.69} & 46.69\scalebox{0.75}{±8.25}         & 66.61\scalebox{0.75}{±1.84}   & 65.05\scalebox{0.75}{±2.07}      & 75.02\scalebox{0.75}{±4.05}   & {72.28\scalebox{0.75}{±3.46}}   &72.74\scalebox{0.75}{±1.51}  &66.38\scalebox{0.75}{±2.48}  &\underline{79.60\scalebox{0.75}{±2.32}}  &78.07\scalebox{0.75}{±3.44}\\ 
GIN   & 57.43\scalebox{0.75}{±8.05} & 43.15\scalebox{0.75}{±6.34} &       70.63\scalebox{0.75}{±1.37}   & 65.80\scalebox{0.75}{±1.34}       & 69.70\scalebox{0.75}{±5.04}   & 55.95\scalebox{0.75}{±6.29}  & 73.63\scalebox{0.75}{±3.78} & 61.97\scalebox{0.75}{±11.53}  &66.35\scalebox{0.75}{±9.16}  &62.59\scalebox{0.75}{±10.24}\\ 
DGCNN   & 57.43\scalebox{0.75}{±3.88} & 41.99\scalebox{0.75}{±7.47}    & 66.65\scalebox{0.75}{±1.55}   & 66.24\scalebox{0.75}{±1.65}      & 63.29\scalebox{0.75}{±3.75}   & 56.11\scalebox{0.75}{±5.30}  &70.58\scalebox{0.75}{±2.84}  &66.07\scalebox{0.75}{±2.33}  &70.45\scalebox{0.75}{±2.86}  &69.95\scalebox{0.75}{±6.61}\\ 
DiffPool    & 57.29\scalebox{0.75}{±5.26} & 47.07\scalebox{0.75}{±6.68}  & 66.78\scalebox{0.75}{±1.98}   & 66.99\scalebox{0.75}{±2.03}         & 66.94\scalebox{0.75}{±4.50}   & 66.32\scalebox{0.75}{±4.19}  &70.69\scalebox{0.75}{±2.58}  &65.69\scalebox{0.75}{±1.66}  &74.95\scalebox{0.75}{±3.05}  &74.52\scalebox{0.75}{±6.11}\\ 
ASAPool    & {58.00\scalebox{0.75}{±5.24}} & 46.53\scalebox{0.75}{±8.33}        & 66.10\scalebox{0.75}{±1.07}   & 65.07\scalebox{0.75}{±2.23}       & 71.31\scalebox{0.75}{±4.04}   & 68.21\scalebox{0.75}{±2.55} &72.47\scalebox{0.75}{±1.67}  &62.36\scalebox{0.75}{±3.02}  &75.75\scalebox{0.75}{±5.77}  &71.04\scalebox{0.75}{±5.64} \\ 
G\textsuperscript{2}GNN   & {56.86\scalebox{0.75}{±7.79}} & 52.56\scalebox{0.75}{±9.81}       & \underline{71.54\scalebox{0.75}{±4.78}}   & 64.08\scalebox{0.75}{±2.11}        & 71.22\scalebox{0.75}{±3.69}   & 69.27\scalebox{0.75}{±4.78}  &65.54\scalebox{0.75}{±1.92}  &64.35\scalebox{0.75}{±2.02} &78.10\scalebox{0.75}{±1.23}  &\underline{80.97\scalebox{0.75}{±1.53}} \\
\midrule
SOLT-GNN    & 56.47\scalebox{0.75}{±3.83} & \underline{53.85\scalebox{0.75}{±6.58}}     & {71.34\scalebox{0.75}{±1.06}}   & \textbf{67.96\scalebox{0.75}{±2.16}}      & {75.20\scalebox{0.75}{±1.76}}   & 72.07\scalebox{0.75}{±3.07}  &\textbf{74.23\scalebox{0.75}{±2.24}}  &\underline{71.67\scalebox{0.75}{±2.23}}  &77.20\scalebox{0.75}{±5.09}  &80.43\scalebox{0.75}{±4.89}\\ 
SizeShiftReg   & {57.07\scalebox{0.75}{±0.95}} & 53.34\scalebox{0.75}{±5.56}       & 70.46\scalebox{0.75}{±1.39}   & 66.17\scalebox{0.75}{±2.67}        & 70.42\scalebox{0.75}{±3.46}   & {72.05\scalebox{0.75}{±2.67}}  &72.39\scalebox{0.75}{±2.57}  &69.13\scalebox{0.75}{±2.89} &79.35\scalebox{0.75}{±6.97}  &79.37\scalebox{0.75}{±2.48}\\ 
\midrule
\textbf{\modelname~(Ours)}        & \textbf{59.43\scalebox{0.75}{±5.97}} & \textbf{55.12\scalebox{0.75}{±6.20}}      & \textbf{72.69\scalebox{0.75}{±1.56}}   & \underline{67.04\scalebox{0.75}{±1.17}}     & \textbf{75.95\scalebox{0.75}{±5.38}}   & \textbf{73.21\scalebox{0.75}{±3.09}}    &\underline{74.11\scalebox{0.75}{±4.23}} &\textbf{72.15\scalebox{0.75}{±1.77}}  &\textbf{82.70\scalebox{0.75}{±4.20}}  &\textbf{81.28\scalebox{0.75}{±4.21}}\\ 
\bottomrule
\end{tabular}} 
\vspace{-0.5em}
\label{main ex}
\end{table*}

\subsubsection{Baselines.}
Our \modelname~allows for working flexibly with most neighborhood aggregation based GNN architectures. 
By default, we employ GIN as the base GNN model in our experiments. 
To evaluate the effectiveness of the proposed \modelname~against the state-of-the-art approaches, we include three types of baselines. (1) \textbf{Graph kernel methods}: Shortest-path graph kernel (SP)~~\cite{borgwardt2005shortest} and Graphlet Kernel (GK)~~\cite{shervashidze2009efficient}. 
They generate graph embeddings based on graph substructures with kernel functions. 
(2) \textbf{Graph neural networks}: base GNN models (GCN~~\cite{kipf2017semi} and GIN~~\cite{xu2018powerful}); DGCNN~~\cite{zhang2018end}, DiffPool~~\cite{ying2018hierarchical} and  ASAPool~~\cite{ranjan2020asap}) design graph pooling mechanisms to obtain expressive graph embeddings; G\textsuperscript{2}GNN~~\cite{liu2022graph} recently proposed to address the graph-level class-imbalance issue, utilizing stochastic augmentations to generate more instances and obtaining extra supervision from graphs with similar structural information.
(3) \textbf{Size-aware GNNs}: SOLT-GNN~~\cite{liu2022size} and SizeShiftReg~~\cite{buffelli2022sizeshiftreg}. 
They both consider the graph size effect on graph neural networks.
The most important baseline is SOLT-GNN, which is the only existing work for the size-imbalance issue and enriches the structure-scarce tail graphs by transferring the head graphs’ structural knowledge.
SizeShiftReg considers the problem of size generalization in out-of-distribution settings, which introduces a regularization strategy that can be applied on GNN frameworks to improve the size-generalization performance from smaller to larger graphs.

\subsubsection{Settings and Parameters.} 
We perform graph classification on five size-imbalanced datasets, where the train/val/test split satisfies the ratio of 6:2:2.
For each dataset, we divide graphs into head and tail following the Pareto principle (the 20/80 rule)~\cite{sanders1987pareto}.
Specifically, we designate the 20\% largest graphs as head graphs, and the rest 80\% as tail graphs.
We construct training sets with high size-imbalanced to validate model's performance, which select from the head and tail sets.
We use accuracy and macro-F1 score as the metrics to evaluate the performance, which are widely adopted in graph classification tasks~~\cite{xu2018powerful, ying2018hierarchical, zhang2018end}. 
For the remaining baselines, we use the source codes provided by the authors. 
We tune hyper-parameters for all models individually as: the propagation layers between graphs $l\in \{1,2,3\}$, the number of neighboring graphs $k\in\{1,2,3,4\}$.
For other hyper-parameters of all models, we follow the settings in their original papers.
We set the depth of GNN backbones as 3 layers in all experiments and the representation dimension of all baselines and \modelname~ to 32. 
The belief propagation step $t$ is set to 2 and weight $\lambda$ is set to 0.5.
The $\epsilon_\mathrm{min}$ and $\epsilon_\mathrm{max}$ are the weight correction factors, which are set to 0.5 and 0.75, respectively.
All experiments are repeated for five times, and we report the averaged results with standard deviations.

\begin{table*}[htbp]
\vspace{-0.5em}
\caption{Graph classification results (accuracy (\%)±std) on head and tail graphs in comparison to other size-aware GNNs.}
\vspace{-1em}
\centering
\resizebox{\textwidth}{!}{
\begin{tabular}{l|ccc|ccc|ccc}
\toprule
\multirow{2}{*}{\textbf{Method}} & \multicolumn{3}{c|}{\textbf{PTC-MR}}    & \multicolumn{3}{c|}{\textbf{FRANKENSTEIN}}    & \multicolumn{3}{c}{\textbf{D\&D}}   \\ 
                        & Test                & Head               & Tail             & Test               & Head                & Tail              & Test               & Head                & Tail               \\ 
\midrule
GIN    & 57.43\scalebox{0.75}{±8.05} & 57.68\scalebox{0.75}{±11.30} & 56.39\scalebox{0.75}{±6.96}        & {70.63\scalebox{0.75}{±1.37}}   & {72.32\scalebox{0.75}{±5.81}}  & 70.31\scalebox{0.75}{±1.32}        & {69.70\scalebox{0.75}{±5.04}}   & 84.27\scalebox{0.75}{±4.47}   & {66.31\scalebox{0.75}{±5.38}} \\
SOLT-GNN    & 56.47\scalebox{0.75}{±3.83} & 61.05\scalebox{0.75}{±10.90} & 54.69\scalebox{0.75}{±5.66}        & {71.34\scalebox{0.75}{±1.06}}   & {73.52\scalebox{0.75}{±5.23}}  & 70.34\scalebox{0.75}{±1.04}        & {75.20\scalebox{0.75}{±1.76}}   & 79.26\scalebox{0.75}{±4.87}   & \textbf{{73.94\scalebox{0.75}{±1.98}} } \\ 
SizeShiftReg   & {57.07\scalebox{0.75}{±6.95}} & 62.53\scalebox{0.75}{±7.05} & 56.96\scalebox{0.75}{±7.30}        & 70.46\scalebox{0.75}{±1.39}   & 70.40\scalebox{0.75}{±1.35}  & {70.70\scalebox{0.75}{±2.60}}        & 70.42\scalebox{0.75}{±3.46}   & 83.02\scalebox{0.75}{±6.04}   & 67.27\scalebox{0.75}{±3.44} \\ 
{\modelname~(Ours)}        & \textbf{59.43\scalebox{0.75}{±5.97}} & \textbf{67.77\scalebox{0.75}{±6.20}} & \textbf{57.29\scalebox{0.75}{±7.44}}        & \textbf{72.69\scalebox{0.75}{±1.56}}   & \textbf{{78.57\scalebox{0.75}{±1.37}}}  & \textbf{71.11\scalebox{0.75}{±2.11}}       & \textbf{75.95\scalebox{0.75}{±3.38}}   & \textbf{88.27\scalebox{0.75}{±2.93}}   & {72.73\scalebox{0.75}{±4.66}}  \\ 
\bottomrule
\end{tabular}}
\vspace{-1em}
\label{table3}
\end{table*}

\begin{table*}[htbp]
\vspace{-0.2em}
\caption{Size-imbalanced graph classification using other GNNs as backbones ( accuracy (\%)±std).}
\vspace{-1em}
\centering
\resizebox{\textwidth}{!}{
\begin{tabular}{l|ccc|ccc|ccc}
\toprule
\multirow{2}{*}{\textbf{Method}} & \multicolumn{3}{c|}{\textbf{PTC-MR}}    & \multicolumn{3}{c|}{\textbf{FRANKENSTEIN}}    & \multicolumn{3}{c}{\textbf{D\&D}}   \\ 
                        & Test                & Head               & Tail             & Test               & Head                & Tail              & Test               & Head                & Tail               \\ 
\midrule
GCN    & 54.29\scalebox{0.75}{±4.69} & 63.88\scalebox{0.75}{±15.1} & 52.01\scalebox{0.75}{±3.21}        & 66.61\scalebox{0.75}{±1.84}   & 66.30\scalebox{0.75}{±5.94}  & 66.71\scalebox{0.75}{±1.00}        & 75.02\scalebox{0.75}{±4.05}   & \textbf{89.83\scalebox{0.75}{±1.73}}   & 71.73\scalebox{0.75}{±5.32}   \\ 
\modelname-GCN   & \textbf{59.14\scalebox{0.75}{±5.83}} & \textbf{67.60\scalebox{0.75}{±8.31}} & \textbf{57.03\scalebox{0.75}{±3.07}}        & \textbf{69.24\scalebox{0.75}{±1.59}}   & \textbf{69.48\scalebox{0.75}{±5.31}}   & \textbf{69.18\scalebox{0.75}{±0.98}}         & \textbf{76.37\scalebox{0.75}{±4.04}}   & 88.37\scalebox{0.75}{±3.45}   & \textbf{73.71\scalebox{0.75}{±5.90}}   \\ 
\midrule
GraphSAGE   & 58.00\scalebox{0.75}{±5.16} & 61.69\scalebox{0.75}{±11.8} & 57.27\scalebox{0.75}{±5.72}        & 69.19\scalebox{0.75}{±1.84}   & 73.09\scalebox{0.75}{±5.35}  & 68.34\scalebox{0.75}{±1.27}        & 68.52\scalebox{0.75}{±5.22}   & 85.61\scalebox{0.75}{±4.43}   & 64.48\scalebox{0.75}{±6.39}  \\ 
\modelname-SAGE        & \textbf{59.44\scalebox{0.75}{±3.95}} & \textbf{67.79\scalebox{0.75}{±4.21}} & \textbf{57.29\scalebox{0.75}{±5.46}}        & \textbf{71.89\scalebox{0.75}{±1.87}}   & \textbf{79.52\scalebox{0.75}{±5.31}}  & \textbf{70.09\scalebox{0.75}{±1.05}}        & \textbf{74.26\scalebox{0.75}{±2.90} }  & \textbf{86.12\scalebox{0.75}{±2.57}}   & \textbf{71.12\scalebox{0.75}{±3.95}} \\
\bottomrule
\end{tabular}}
\vspace{-0.3cm}
\label{table4}
\end{table*}

\subsection{Performance Evaluation}
We conduct graph classification on the five size-imbalanced datasets for comparison.
For a fair comparison, we employ GIN as the base model for all the GNN-based approaches.

\subsubsection{\modelname~ for Real-world Graphs.}

The overall accuracy of different baselines is shown in Table~\ref{main ex}.
The best results are shown in bold and the runner-ups are underlined. 
\modelname~ shows superiority in improving the overall performance across different metrics. 

In particular, we make the following observations.
Compared to the pooling approaches such as SortPool, DiffPool, and ASAPool, \modelname~ achieves higher performance with a large margin in all datasets.
Though they focus on the designed pooling mechanisms to enhance the graph representations, they are unable to address great structural discrepancies in the size-imbalanced datasets, thus compromise model performance.
G\textsuperscript{2}GNN connects graphs based on topological similarity, which prevents direct connections between small and large graphs, thereby failing to alleviate the structural differences caused by the size-imbalance issue.
\modelname~ also outperforms approaches SOLT-GNN and SizeShiftReg, which investigate the graph size effect on the performance.
This proves that the construction of a higher-level graph abstraction based on a discriminative graph embedding module contributes to smoothing the structural features between head and tail graphs, as well as re-weighting each graph based on the estimated energy score, which is effective for improving  the model's capacity for generalization.

\begin{figure*}[!htbp]
\vspace{-1em}
\centering
\subfigure[\modelname~ under different levels of size imbalance.]{
\begin{minipage}[t]{ 0.23\linewidth }
\centering
\includegraphics[width=\linewidth]{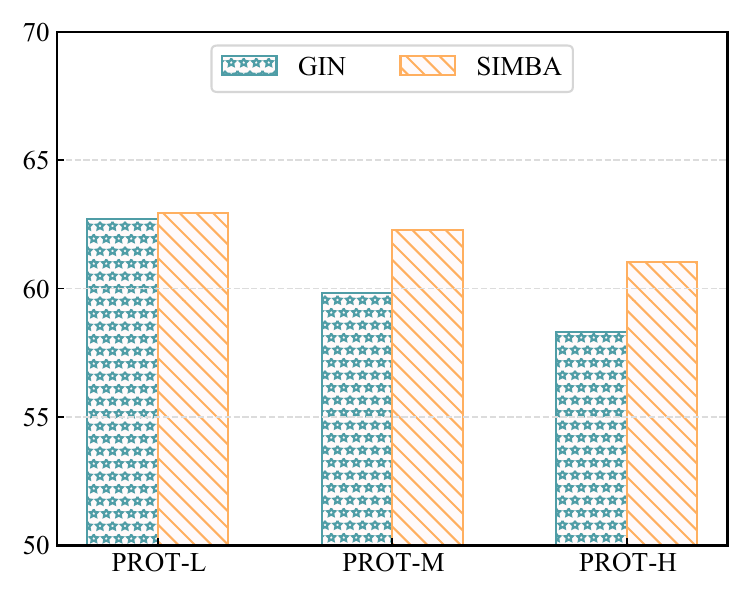}\vspace{-2.5em}
\label{fig: sizeimb}
\end{minipage}
}
\subfigure[Ablation study results for \modelname.]{
\begin{minipage}[t]{ 0.23\linewidth }
\centering
\includegraphics[width=\linewidth]{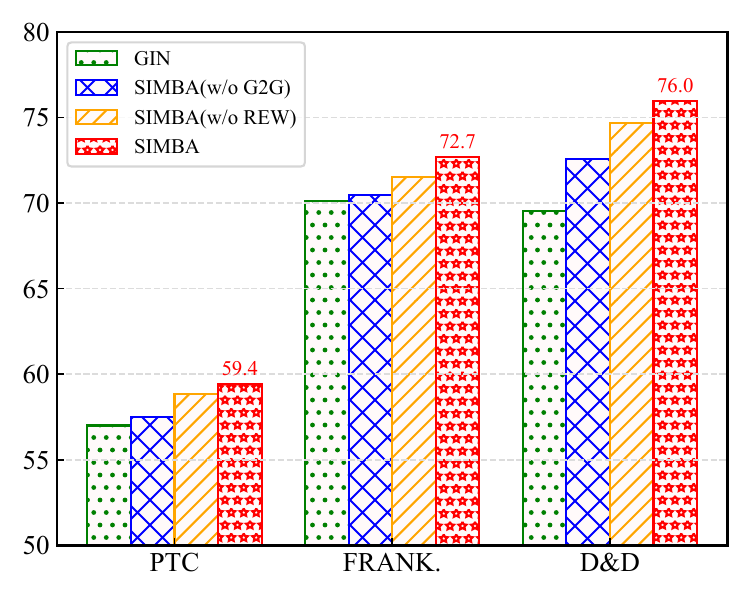}\vspace{-2.5em}
\label{fig: ablation}
\end{minipage}
}
\subfigure[Relationship between neighborhood number and performance on PROTEINS.]{
\begin{minipage}[t]{ 0.23\linewidth}
\centering
\includegraphics[width=\linewidth]{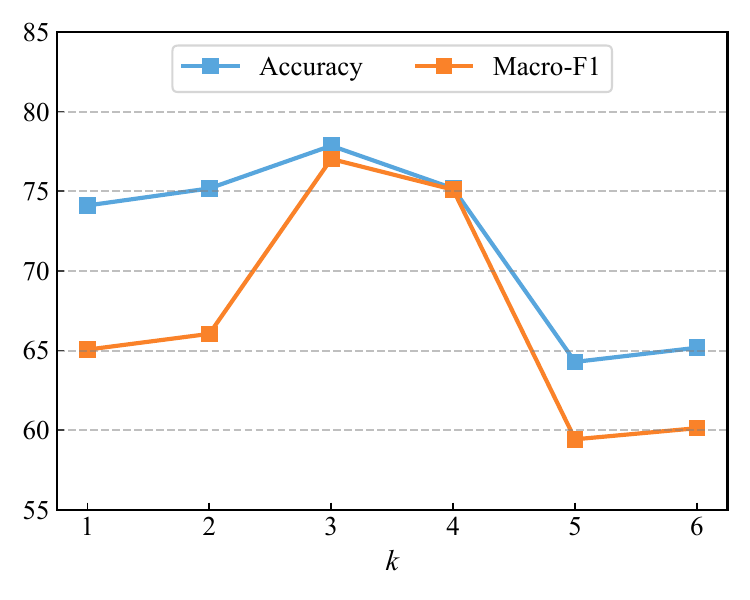}\vspace{-2.5em}
\label{fig:T=10}
\label{fig: protein}
\end{minipage}
}
\subfigure[Relationship between neighborhood number and performance on D\&D.]{
\begin{minipage}[t]{ 0.23\linewidth}
\centering
\includegraphics[width=\linewidth]{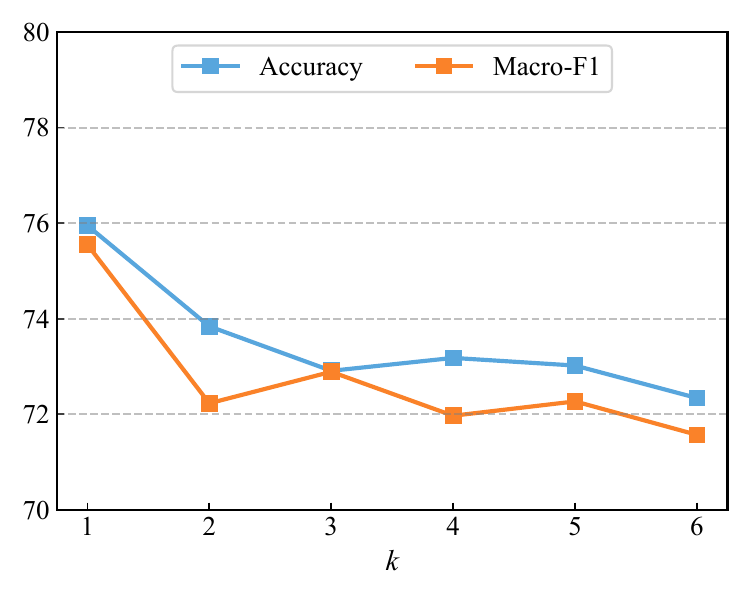}\vspace{-2.5em}
\label{fig: dd}
\end{minipage}
}
\centering
\caption{Performance analysis of \modelname. }
\vspace{-1.5em}
\label{fig:exp}
\Description{}
\end{figure*}
\subsubsection{\modelname~ with Size-aware GNNs.}
To further analyze the generalization to both large and small graphs, we illustrate the performance promotion from three perspectives, i.e., Test (the entire accuracy of the test set), Head and Tail (the head and tail graphs' accuracy of the test set) on three datasets.
Note that, SOLT-GNN~~\cite{liu2022size} achieves fewer improvements on head graphs in all datasets, and its performance on D\&D is inferior to that of GIN.
This phenomenon could result from that it unintentionally create an artificial out-of-distribution scenario, where they only observe tail graphs during the training, leading to downgraded performance for head graphs that GNNs originally perform well at.
Another size-aware baseline SizeShiftReg investigates the size-generalization capabilities from smaller to larger graphs and still suffers from the considerable discrepancies of structural features in the training progress, which hinders the generalization capability in size-imbalanced datasets.
Comparing \modelname~ with baselines, we notice that our proposed \modelname~ improves the performance of both head and tail graphs.
Bridging head and tail graphs with similar emdeddings provide extra information for each graph in graphs-to-graph. 
Re-weighting the incompatible graphs among neighboring graphs  smooths structural discrepancies and further boost overall performance.

\subsubsection{\modelname~ with other GNNs Backbones.}
To demonstrate the proposed \modelname's flexibility on other GNN backbones, we verify the \modelname~ using GCN and GraphSAGE as the backbones for evaluation. 
We show the classification accuracy scores in Table \ref{table4}, where the best results are shown in bold. 
We can observe that, our \modelname~ can outperform corresponding GNN backbones in each case, illustrating the flexibility of \modelname~to apply with different GNN architectures and enhance the generalization capacity to head and tail graphs.

\subsubsection{\modelname~ under Different Levels of Size Imbalance.} 
To further analyze \modelname~'s ability in alleviating the size-imbalanced issue, we create training sets with different size-imbalanced ratio (SIR) on PROTEINS, as defined in~\ref{definition}.
We create 10\%/10\% splits for the training/validation sets, ensuring that the training and validation sets are class-balanced.
The remaining graphs are used as the testing set.
The SIRs depend on the proportion of graphs selected from the head and tail sets. 
We repeat the graph selection process multiple times, calculate the resulting imbalance ratios, and choose three groups of splits exhibiting significant variations in SIR.  
These groups are denoted as PROT-L (SIR=$\log_{2}{12}$), PROT-M (SIR=$\log_{2}{25}$), and PROT-H (SIR=$\log_{2}{54}$).
Note that the higher SIR means higher size-imbalanced distribution in the dataset.
We evaluate \modelname~ with GIN as the backbone and show the dataset information, the Accuracy scores in Figure \ref{fig: sizeimb}.
The performance of graph representation learning generally gets worse with the increase of SIR of the dataset. 
\modelname~ performs better under all different SIR scenarios and consistently boost performance by a large margin.
The two modules (graphs-to-graph propagation and energy-based graph re-weighting) in \modelname~ mitigate the structural discrepancies and improve the model's generalization capacity for size-imbalanced issue.

\subsection{Analysis of \modelname~}

\subsubsection{Ablation Study.}
We conduct ablation variant studies for the main mechanisms of \modelname~ to evaluate the contribution of each component.
The variants include: 
\begin{itemize}[leftmargin=1.5em]
    \item \textbf{\modelname(w/o G2G)}, which removes the graphs-to-graph propagation (Section \ref{GoG}) and energy-based graph re-weighting mechanism (Section \ref{reweight}) of \modelname, employs the discriminative graph embedding module for graph representation learning in Eq.\ref{eq:pool};
    \item \textbf{\modelname(w/o Rew)}, which removes the energy-based graph re-weighting mechanism (Section \ref{reweight}) of \modelname.
\end{itemize}

In Figure \ref{fig: ablation}, we present the performance results (accuracy) of the ablation study.
The graphs-to-graph propagation module contributes the most, which can achieve at most 6.9\% improvement in terms of Accuracy score with only the inherent structural information of graphs.
When we remove the  graphs-to-graph propagation module, complementary information from neighboring graphs cannot be effectively passed between graphs, resulting in inferior performance.
Without the energy-based graph re-weighting operation on each instance, the performance also generally decreases.

\subsubsection{Hyper-parameter Analysis.}
In Figure \ref{fig: protein} and \ref{fig: dd}, we analyze the influence of the number of neighboring graphs on the performance of \modelname~ on PROTEINS and D\&D. 
The experimental setting is the same as Section 5.1.3, except that we vary the $k$ among $\{1, ..., 6\}$.  
We use Accuracy and Macro-F1 as evaluation metrics to evaluate the performance of ~\modelname.
The optimal number of neighboring graphs for achieving the best performance varies across datasets.
In PROTEINS, the performance increases first as $k$ increases to 3  since higher $k$ means more number of neighboring graphs connect, which smooths structural feature discrepancies between more head and tail graphs in the same category.
However, when $k$ proceeds to increase, the overall performance exhibits a declining trend, which is also observed in the D\&D.
This is attributed to the fact that more added neighborhoods may aggregate over-smoothing features.  

\setlength{\parskip}{5pt}
\section{CONCLUSION}

In this paper, we propose a novel framework named \modelname~ for the size-imbalanced graph classification issue.\
Specifically, \modelname~ first introduce a discriminative graph embedding technique that facilitates the derivation of a fixed-length embedding vector for each variable-sized graph.
After obtaining the graphs' embeddings, \modelname~ constructs a higher-level abstraction (Graphs-to-Graph) to smooths the features among head and tail graphs with the same category.
Then \modelname~re-weights the graphs via the energy-based belief propagation mechanism, aggregating energy scores in the neighborhoods and adjusting the influence of each graph to further smooth the structural discrepancies in the local region during the training process.
Comprehensive experiments on multiple real-word benchmark datasets demonstrate the effectiveness and the generalization capacity of \modelname.

\begin{ack}
The corresponding author is Jianxin Li. 
This work was supported by the National Natural Science Foundation of China through grants 62225202, 62302023 and 62462007.
\end{ack}

\clearpage
\begin{ethic}
Considering ethical standards, all datasets were obtained from publicly accessible repositories with clear permissions for research use. For data derived from user-generated content or social platforms, we adhere to platform terms that specify consent for research purposes. We strongly encourage users to approach the dataset responsibly, remaining mindful of its ethical implications.
\end{ethic}

\bibliographystyle{wsdm2025}
\balance
\bibliography{wsdm2025}
\clearpage

\end{document}